%% file: main.tex
\documentclass[letterpaper]{article} 
\usepackage{aaai25}  
\usepackage{times}  
\usepackage{helvet}  
\usepackage{courier}  
\usepackage[hyphens]{url}  
\usepackage{graphicx} 
\usepackage{natbib}  
\usepackage{caption} 
\usepackage{algorithm}
\usepackage{algorithmic}

\usepackage{newfloat}
\usepackage{listings}
\DeclareCaptionStyle{ruled}{labelfont=normalfont,labelsep=colon,strut=off} 
\floatstyle{ruled}
\newfloat{listing}{tb}{lst}{}
\floatname{listing}{Listing}
\usepackage{amsfonts}
\usepackage{amsmath}
\usepackage{booktabs}
\usepackage{threeparttable}
\usepackage{subfigure}
\usepackage{xspace,tabularx,multirow}
\usepackage{tikz}
\usepackage{pgfplots}

\usepackage{esvect}
\usepackage{enumitem}

\usepackage{stmaryrd}

\include{cmd.tex}

\title{\algo: A Ring-Enhanced Graph Transformer for Organic Solar Cell Property Prediction}
\author{
    Zhihao Ding\textsuperscript{\rm 1}\equalcontrib, Ting Zhang\textsuperscript{\rm 1}\equalcontrib, Yiran Li\textsuperscript{\rm 1}, Jieming Shi\textsuperscript{\rm 1}, Chen Jason Zhang\textsuperscript{\rm 1}
}
\affiliations{
    \textsuperscript{\rm 1}Department of Computing, The Hong Kong Polytechnic University, Hong Kong SAR\\
    \{tommy-zh.ding, leyla2.zhang, yi-ran.li\}@connect.polyu.hk, \{jieming.shi, jason-c.zhang\}@polyu.edu.hk
}

\begin{document}

\maketitle

\begin{abstract}
Organic Solar Cells (OSCs) are a promising technology for sustainable energy production.  However, the identification of molecules with desired OSC properties typically involves laborious experimental research. To accelerate progress in the field, it is crucial to develop machine learning models capable of accurately predicting the properties of OSC molecules.  While graph representation learning has demonstrated success in molecular property prediction, it remains under-explored for OSC-specific tasks.
Existing methods fail to capture the unique structural features of OSC molecules, particularly the intricate ring systems that critically influence OSC properties, leading to suboptimal performance.  To fill the gap, we present \algo, a novel graph transformer framework specially designed to capture both atom and ring-level structural patterns in OSC molecules. \algo constructs a hierarchical graph that integrates atomic and ring structures and employs a combination of local message-passing and global attention mechanisms to generate expressive graph representations for accurate OSC property prediction.
We evaluate \algo's effectiveness on five curated OSC molecule datasets through extensive experiments. The results demonstrate that \algo consistently outperforms existing methods, achieving a 22.77\% relative improvement over the nearest competitor on the \cepdb dataset.
\end{abstract}

%
\begin{links}
    \link{Code}{https://github.com/TommyDzh/RingFormer}
\end{links}

\input{introduction}

\input{relatedworks}

\input{preliminary}

\input{method}
\input{experiments}

\input{conclusion}

\bibliography{references}
\input{appendix}

\end{document}

%% file: cmd.tex
\newcommand{\etal}{{\it et al.}\xspace}

\def\header{\vspace{1pt} \noindent}

\newcommand{\algo}{RingFormer\xspace}

\newcommand{\graph}{hierarchical OSC graph\xspace}

\newcommand{\ga}{atom-level graph\xspace}
\newcommand{\gr}{ring-level graph\xspace}
\newcommand{\gi}{inter-level graph\xspace}

\newcommand{\atom}{atom-level message passing module\xspace}

\newcommand{\ring}{ring-level cross-attention module\xspace}

\newcommand{\inter}{inter-level message passing module\xspace}

\newcommand{\cepdb}{CEPDB\xspace}
\newcommand{\hopv}{HOPV\xspace}
\newcommand{\polymer}{PFD\xspace}
\newcommand{\nnfa}{NFA\xspace}
\newcommand{\pnfa}{PD\xspace}

\newcommand{\maccs}{MACCS\xspace}
\newcommand{\ecfp}{ECFP\xspace}
\newcommand{\gine}{GINE\xspace}
\newcommand{\gvn}{GINE-VN\xspace}
\newcommand{\afp}{AttentiveFP\xspace}
\newcommand{\ognn}{O-GNN\xspace}
\newcommand{\topk}{TopKPool\xspace}
\newcommand{\sag}{SAGPool\xspace}

\newcommand{\gps}{GPS\xspace}
\newcommand{\graphormer}{Graphormer\xspace}
\newcommand{\vit}{GraphViT\xspace}
\newcommand{\transformer}{Vanilla Transformer\xspace}

\newcommand{\voc}{$V_{oc}$\xspace}
\newcommand{\jsc}{$J_{sc}$\xspace}

\newcommand{\G}{\mathcal{G}\xspace}
\newcommand{\GA}{G_{A}\xspace}
\newcommand{\GR}{G_{R}\xspace}
\newcommand{\GI}{G_{I}\xspace}

%% file: introduction.tex
\section{Introduction}\label{sec:introduction}
\vspace{2mm}

As the demand for renewable energy sources grows, organic solar cells (OSCs) have attracted considerable interest for their ability to convert sunlight into electricity~\cite{wang2016difluorobenzothiadiazole}. Despite their potential, the development of OSCs has been hindered by the reliance on trial-and-error methods, which involve complex and time-consuming synthesis procedures~\cite{sun2019use}.
To accelerate progress, there is increasing interest in leveraging machine learning models to accurately predict the properties of OSC molecules, promising  to expedite the development of OSCs.

\begin{figure}
    \centering
      \includegraphics[width=0.49\textwidth]{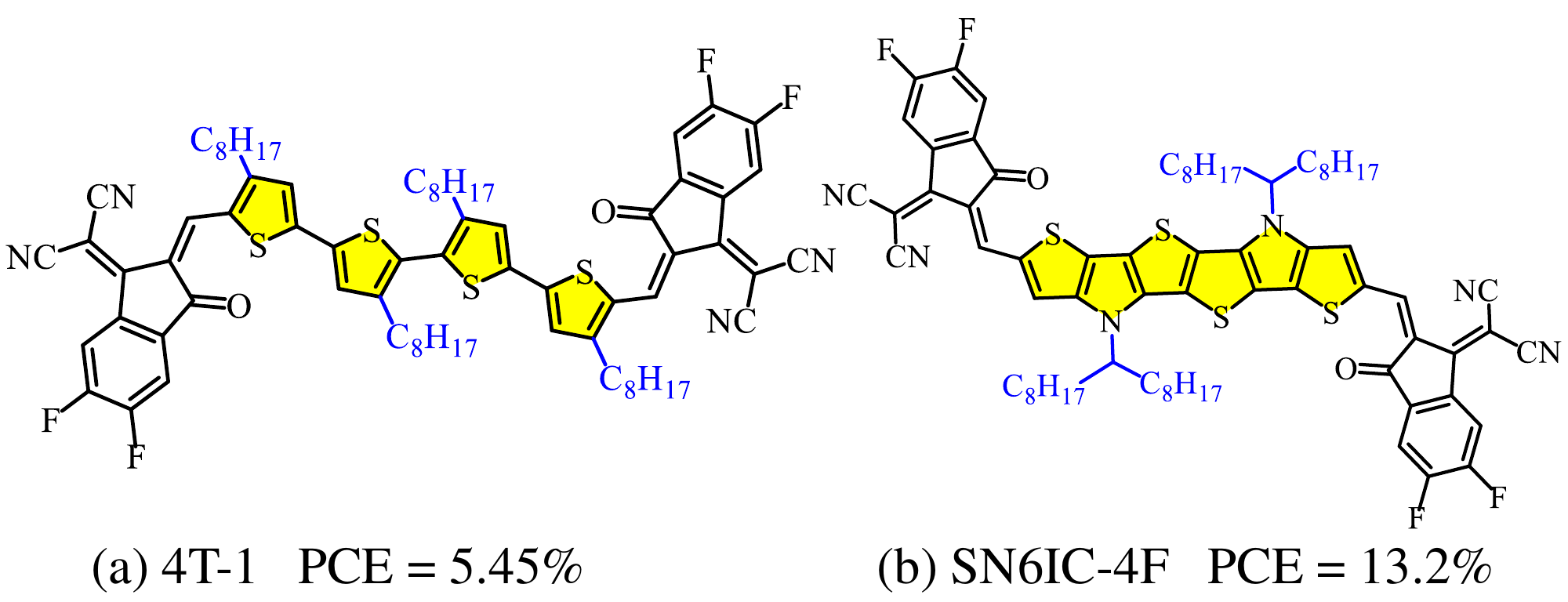}
    
    \vspace{-2mm}
    \caption{Example of OSC molecules.}
    \label{fig:molecules}

\vspace{-8mm}
\end{figure}

In this work, we aim to predict OSC properties based on the structure of OSC molecules, i.e., organic small-molecule semi-conducting materials for the active layer of OSCs. Such molecules function as electron acceptors or donors to create photovoltaic effects with efficacy linked to their conjugated structure, such as aromatic rings~\cite{solakAdvancesOrganicPhotovoltaic2023}. In chemistry, rings are closed loops of atoms connected through covalent bonds~\cite{clayden2012organic}, and the design of complex ring systems has become a central focus in OSC research~\cite{schweda2021recent}. These systems can include various fused and non-fused ring structures~\cite{gaoRecentProgressNonfused2023}.
For example, as shown in Figure~\ref{fig:molecules}, the core of OSC molecule 4T-1 (highlighted in yellow) on the left comprises four non-fused thiophene rings connected by single bonds. In contrast, the core of SN6IC-4F on the right side is a fused S,N-heteroacene consisting of six rings, where adjacent rings share a pair of bonded atoms.
The backbone of both molecules can be characterized by 
multiple inter-connected rings connecting  each other in various ways, forming a complex ring system.
Additionally, different alkyl functional groups (in blue) attached to the core as side chains can further influence OSC performance~\cite{chingImpactAlkylBasedSide2023}. 
These structural differences ultimately lead to significantly different power conversion efficiency (PCE) values between 4T-1 and SN6IC-4F.
Therefore, accurately predicting the properties of OSC molecules requires capturing both the high-level structure of the ring system and the local-level atomic groups within these molecules.

For predicting molecular properties, graph neural networks (GNNs) have been widely adopted, they represent chemical molecules as graphs, with nodes for atoms and edges for chemical bonds.
While GNNs effectively capture local atomic structures like functional groups, they struggle to model higher-order patterns such as those found in OSC ring systems~\cite{he2023generalization}. To address these limitations, graph pooling methods~\cite{gao2019graph,lee2019self} and motif-based approaches~\cite{yu2022molecular,zang2023hierarchical} have been developed to extend graph learning beyond localized atomic features by leveraging substructures and motifs. However, they fail to emphasize the rings and their interconnections which are crucial for OSC molecules. 
This limitation is particularly problematic as GNNs are inherently weak in modeling ring-like structures and long-range dependencies~\cite{loukas2019graph,chen2020can,rampavsek2022recipe}, leading to suboptimal performance in OSC property prediction~\cite{eibeck2021predicting}.
Recently, Zhu~\etal ~\cite{zhu2023o} enhanced GNNs by incorporating additional ring representations, but they still neglect the crucial interconnections between rings.
Thus, current graph-based models are inadequate for effectively highlighting the ring systems in OSC molecules, thereby limiting their predictive accuracy for OSC properties.

To fill the gap, we introduce \algo, a novel graph transformer framework designed to capture structural patterns at both the atom and ring levels within OSC molecules. As illustrated in Figure \ref{fig:frame}, \algo firstly constructs a \graph to explicitly model atom groups and the ring system in an OSC molecule. 
The \graph comprises three levels: the atom-level graph, the ring-level graph, and the inter-level graph. The atom-level graph describes the atomic bonding structure of the OSC molecule. Above this, the ring-level graph focuses on the rings and their interconnections, capturing the high-level ring system. Finally, the inter-level graph models the relationships between rings and atoms, representing the hierarchical structure of the molecule.
By integrating these three levels, the \graph provides a comprehensive depiction of OSC molecular structures, enabling more accurate prediction of OSC properties.

Subsequently, \algo combines the power of local message-passing and global attention to capture the distinct structural patterns in each level and learn expressive graph representations. 
On atom-level graphs, \algo layers use message-passing GNNs to encode localized structures into atom node representations.  For ring-level graphs, \algo layers incorporate a novel cross-attention mechanism to capture global patterns within the ring system. Specifically, the proposed mechanism effectively and efficiently captures the interconnections between rings.
\algo layers further facilitate interactions between ring and atom nodes through message-passing on the inter-level graph.
 At the end of each \algo layer, a hierarchical fusion strategy is implemented to allow information from different levels to complement each other.
After stacking multiple layers, \algo aggregates both atom and ring node representations into graph representations that comprehensively encode the structure of OSC molecules. To evaluate the effectiveness of \algo in OSC property prediction, we compare 
 \algo against 11 baselines over 5 OSC molecule datasets.
Experimental results show that \algo outperforms baselines consistently. Notably, on the large-scale CEPDB dataset, \algo achieves a remarkable 22.77\% relative improvement over the closest competitor.  In summary, our contributions are: 

 \begin{itemize}[leftmargin=*]
     \item We study the important problem of OSC property prediction and present \algo, the first graph transformer framework capturing ring systems within OSC molecules.
     \item We construct a \graph to depict OSC molecule at atom and ring levels. Message-passing and transformer-based attention mechanism are combined to learn expressive representations on the hierarchical graph. 
     
     \item We further capture the inter-level interactions between rings and atoms in the hierarchical graph, and design a hierarchical fusion strategy to strength the quality of learned representations.
     \item Extensive experiments on 5  OSC property prediction datasets validate the superior performance of \algo.
 \end{itemize}

%% file: relatedworks.tex
\section{Related Work}\label{sec:related}
\vspace{2mm}
\header
\textbf{OSC Property Prediction.}
Organic solar cells (OSCs) have garnered significant research attention as one of the most promising technologies for harnessing solar energy~\cite{eibeck2021predicting}. As conducting laboratory experiments to screen candidate OSC molecules is time and resource-intensive~\cite{xu2022polymer}, researchers have recently turned to machine learning methods for efficient OSC property prediction.
Currently, fingerprint-based approaches~\cite{eibeck2021predicting} are commonly employed. Typically, these methods utilize hand-crafted molecular fingerprints such as MACCS~\cite{durant2002reoptimization} and ECFP~\cite{rogers2010extended} as molecular features, which are then inputted into off-the-shelf machine learning models like random forest and support vector machine. However, fingerprints represent simplified abstractions of molecular structures, which overlook crucial molecular information and interactions, particularly in OSC molecules with complex structures~\cite{miyake2021machine}. Inspired by the success of GNNs in drug discovery, Eibeck \etal~\cite{eibeck2021predicting} recently explored the application of GNNs in OSC property prediction. However, they found that conventional GNNs often perform poorly in predicting OSC properties, achieving lower accuracy compared to fingerprint methods. The development of effective models for OSC property prediction remains under-explored.

\header
\textbf{Graph Representation Learning on Molecules.}
As molecules can be naturally represented as graphs, graph neural networks (GNNs) ~\cite{hu2019strategies,zhu2023o} are widely used for molecule property prediction. 
However, conventional GNNs struggles to capture high-order structures~\cite{he2023generalization}, including important molecular features such as rings~\cite{chen2020can,loukas2019graph}. To address the limitations of GNNs, researchers have developed graph pooling methods~\cite{gao2019graph,lee2019self}, motif-based methods~\cite{yu2022molecular,zang2023hierarchical}, and graph transformers~\cite{rong2020self,ying2021transformers,kim2022substructure}. Although these models capture higher-level molecular patterns beyond localized atomic features, they still fail to emphasize the rings and their connections, which are crucial for accurately predicting OSC molecule properties.
To fill the gap, 
we propose \algo, the first graph transformer framework that is specially designed to capture ring systems for OSC property prediction.

%% file: preliminary.tex
\begin{figure*}[t]
	\centering
		\includegraphics[width=0.88\textwidth]{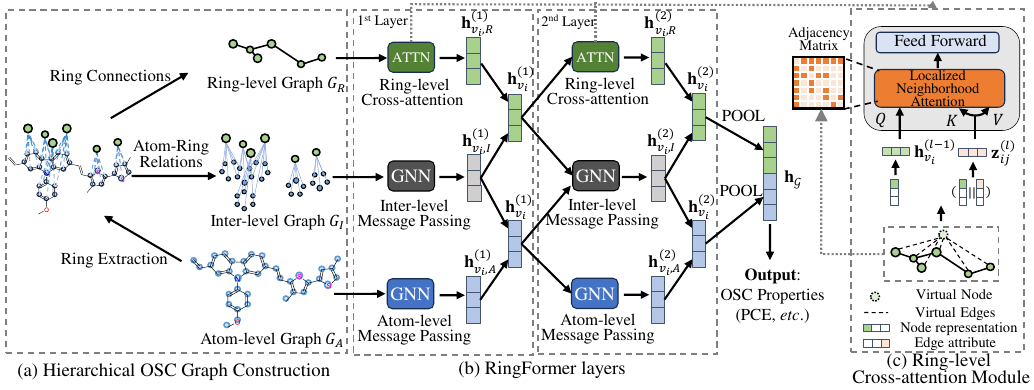}
   \vspace{-3mm}
 \caption{The \algo framework. For clarity, we showcase the framework with $L=2$ \algo layers.}
\label{fig:frame}   
\vspace{-4mm}
\end{figure*}

\section{Problem Formulation}
\label{sec:definition}

\noindent \textbf{Data Model.}
Following previous works on OSC property prediction~\cite{eibeck2021predicting}, an OSC molecule is represented by an atom-level molecular graph $\GA=(V_{A}, E_{A})$, where $V_{A}$ is the set of nodes representing atoms and $E_{A}$ is the set of edges representing chemical bonds. Let $\mathbf{x}_{v_{i}} \in \mathbb{R}^{d_{V_A}}$ denote node attribute vector of node $v_{i} \in V_{A}$ and  $\mathbf{e}_{ij}  \in \mathbb{R}^{d_{E_A}}$  denote edge attribute vector of edge $e_{ij}\in E$, where $d_{V_A}$ and $d_{E_A}$ are the dimension of node and edge attributes, respectively. 

\header\textbf{Prediction Task.} Given an OSC molecular graph $\GA$, the objective of our work is to learn a property prediction model $f:\GA\rightarrow y$ that predicts the target property value $y$, such as power conversion efficiency (PCE), highest occupied molecular orbital (HOMO), lowest unoccupied molecular orbital (LUMO), etc~\cite{miyake2021machine}. Notably, PCE holds particular significance for OSC molecules, as it serves as a pivotal indicator of their efficiency in converting sunlight into electrical energy~\cite{eibeck2021predicting}.  Accurate prediction of PCE is essential  for assessing the performance of OSCs. In this study, we prioritize the prediction of PCE as our pivotal target property.

%% file: method.tex
\section{The \algo Method}\label{sec:method}

\noindent \textbf{Overview.} 
Aiming to capture both atom and ring-level structural patterns in an OSC molecule, our proposed \algo framework firsts constructs a \graph, and  
then holistically encodes the hierarchical graph using \algo layers for property prediction. As depicted in Figure~\ref{fig:frame}, the \graph consists of atom, ring, and inter-level graphs that models OSC molecule structure from multiple levels.
The \algo layers combine message-passing and global attention mechanisms to further 
learn node representations at each level and fuse information across hierarchies. Finally, \algo aggregates node representations of both atom and ring nodes to generate the graph representation  which is used for predicting target OSC properties.

\subsection{Hierarchical OSC Graph Construction}\label{sec::construction}
In this section, we introduce how to construct the \graph  $\mathcal{G}$
for an OSC molecule. As illustrated in Figure \ref{fig:molecules}, the ring system, characterized by the types of 
rings and their interconnections, plays a crucial role in determining the properties of OSC molecules. To overcome the limitations of a flat molecular graph $\GA$ that solely depicts low-level atom-based structure, our proposed \graph  $\mathcal{G}$
comprehensively represents an OSC molecule from three levels. As shown in Figure \ref{fig:frame}, given an input \ga $\GA$,
we construct a \gr $\GR$ above the atom-level $\GA$ to explicitly depict the high-level ring system. 
After having $\GA$ and $\GR$, we further incorporate a bipartite \gi $\GI$ to describe the connections between atom and ring nodes and represent graph hierarchy.  Combining the three levels of graphs, we obtain the \graph $\G=\{\GA, \GR, \GI\}$. Next, we provide more details on each level of the graph.

\header\textbf{Ring-level Graph Construction.}
Given an input OSC molecule modeled by an atom-level molecular graph $\GA$ (introduced in the problem formulaiton section), we first extract the smallest rings from $\GA$ and represent each ring by a ring node in $V_{R}=\{v_{1}, v_{2},\cdots\,v_{|V_{R}|}\}$. Here, the smallest ring is defined as a ring such that no proper subset of its nodes can form a smaller ring.

Different types of rings are characterized by their atom composition. For a ring node $v_{i}\in V_{R}$, we use the one-hot encoding of its ring type as the attribute vector $\mathbf{x}_{v_{i}} \in \mathbb{R}^{d_{V_R}}$, where $d_{V_R}$ represents the total number of ring types. Between a pair of extracted ring nodes, we add an edge in the edge set $E_{R}$ to suggest the connections between them when (1) the two rings share one or more atoms or (2) the two rings are connected by a single chain of one or more non-aromatic bonds.
For an edge $e_{ij} \in E_{R}$, its edge attribute   is an one-hot vector $\mathbf{e}_{ij} \in \mathbb{R}^{d_{E_R}}$ indicating the connection type, where $d_{E_R}$ represents the total number of connection types.
We differentiate the type of connection between two ring nodes based on the atoms shared by two rings in condition (1) or the atoms and bonds composition of the connecting chain in condition (2). Having the ring nodes and their interconnections, we construct \gr $\GR=(V_{R}, E_{R})$.

\header\textbf{Inter-level Graph Construction.} We further construct an \gi $\GI$ to make $\GA$ and $\GR$ interact with each other and depict graph hierarchy.
Specifically, for a ring node $v_{i} \in V_{R}$ and an atom node $v_{j} \in V_{A}$, we connect them using   $e_{ij} \in E_{I}$ if the atom node is one of the constituents of the ring, represented by the ring node.
In this way, we model the graph hierarchy based on the  membership between atom and ring nodes in the \gi and have $\GI=(V_{A}\cup V_{R}, E_{I})$. Note that $\GI$ is an undirected graph, enabling bidirectional message propagation between any pair of connected atom and ring nodes.

Finally, we combine the three levels of graphs as 
the \graph  $\G=\{\GA, \GR, \GI\}$
which comprehensively encapsulates the complex structure of OSC molecules, 
enabling accurate property prediction.

\subsection{\algo Layer}\label{sec::layer}
In this section, we present \algo layers that learn expressive representations for both atom and ring nodes on the \graph $\mathcal{G}$. Considering the hierarchical nature of $\mathcal{G}$, we combine the power of message-passing and global attention in \algo layers to better capture the unique patterns in each graph level. As shown in Figure~\ref{fig:frame}, each \algo layer consists of an \atom, a \ring, and an \inter. At the end of each \algo layer, we fuse the latent information learned from hierarchical levels to generate expressive node representations for atoms and rings, which are then forwarded to the next layer for further updates. Next, we introduce each module in detail.

\header\textbf{Atom-level Message Passing Module.}
On the \ga $\GA$, we aim to learn atom node representations that capture the local chemical structures in OSC molecules, such as function groups. As GNNs are good at mining patterns in local structures~\cite{he2023generalization},  we adopt GNNs to perform message passing on $\GA$. In the $l$-th \algo layer, GNNs are applied on the \ga as follows:
\begin{equation}\label{eq:gnn_a}
\small
    \mathbf{h}^{(l)}_{v_{i},A} = \mathrm{GNN}_{A}^{(l)}(\mathbf{h}^{(l-1)}_{v_{i}}, \{\mathbf{h}^{(l-1)}_{v_{j}}|v_{j}\in \mathcal{N}_{\GA}(v_{i})\}, \{\mathbf{e}_{ij}\}),
\end{equation}
where $v_{i},v_{j} \in V_{A}$ denote nodes in $\GA$, $\mathcal{N}_{\GA}(v_{i})$ is the neighborhood of node $v_{i}$ in $\GA$, $\mathbf{e}_{ij}$ is the edge attribute of edge $e_{ij} \in E_{A}$, and $\mathbf{h}^{(l-1)}_{v_{i}} \in \mathbb{R}^{d}$ is the node representation of node $v_{i}$ from the preceding \algo layer. Specially, we have $\mathbf{h}^{(0)}_{v_{i}} =  \mathbf{W}_{a}\mathbf{x}_{v_{i}}$, where $\mathbf{W}_{a}\in \mathbb{R}^{d \times d_{V_A}}$ are learnable parameters to transform node attribute $\mathbf{x}_{v_{i}}$, $d$ is the hidden dimension. Here, $\mathrm{GNN}_{A}^{(l)}$  can be arbitrary GNN architecture. Throughout the paper, we use GINE~\cite{hu2019strategies} as the default GNN backbone. In Appendix, we study the effect of different GNN backbones. $\mathbf{h}^{(l)}_{v_{i},A}\in \mathbb{R}^{d}$ is the output node representation of the \atom in the $l$-th \algo layer.

\header\textbf{Ring-level Cross-attention Module.} 
In the encoding of \gr,  we aim to learn ring node representations that capture the global structure of the ring system. To achieve this, we introduce a novel transformer-based cross-attention module tailored for the \gr. 
Recently, graph transformers~\cite{ying2021transformers,kreuzer2021rethinking} have shown their superiority to GNNs in modeling long-range node dependencies and capturing structural patterns beyond localized graph structures~\cite{rampavsek2022recipe}.

However, existing graph transformers still exhibit limitations in capturing the chemical semantic-rich structures of  ring-level graphs that represent ring systems in OSC molecules. 
A primary limitation of existing graph transformers is their inadequate utilization of edge attributes which contain crucial information about how two rings are connected in the \gr. 
The graph transformers typically compute self-attention using node representations as queries, keys, and values, with edge attributes merely considered in attention biases. Consequently, valuable chemical semantics in edge attributes of the \gr are not integrated into the output ring node representations. 

Moreover, graph transformers further suffer from efficiency issues when integrating edges attributes. For instance, as Graphormer
calculate fully-connected attention among all node pairs in a graph, it also needs to compute the shortest paths among all node pairs  as edge attributes, leading to high computation complexity relative to the number of nodes~\cite{shirzad2023exphormer}.

To address the above limitations, we introduce two key designs in our proposed \ring: cross-attention mechanism and localized neighborhood attention strategy. These designs optimize the utilization of edge attributes, while maintaining computational efficiency. Firstly, to better leverage the edge attributes, we propose to adopt a \emph{cross-attention} strategy instead of the node-centric \emph{self-attention} commonly used in graph transformers. 
Specifically, for a source ring node $v_{j} \in V_{R}$ connected to the target ring node $v_{i} \in V_{R}$ through edge $e_{ij} \in E_{R}$, we regard the node representation of $v_{i}$ as the query, while the combination of the node representation of $v_{j}$ and edge attribute $\mathbf{e}_{ij}$ serves as the key and value. 
In such a way, edge attributes are not only considered in the attention scores but also integrated into the target node representation during aggregation. 
Then, in calculating multi-head attention, the queries, keys, and values for the $c$-th attention head  in the $l$-th layer are calculated as follows: 
\begin{equation}\label{eq:tokens}
\small
\begin{split}
\mathbf{z}_{ij}^{(l)} &= \mathrm{MLP}_{Z}^{(l)} (\mathbf{h}_{v_{j}}^{(l-1)}||\mathbf{e}_{ij}), \\
    \mathbf{q}_{i,c}^{(l)}= \mathbf{W}_{q,c}^{(l)} \mathbf{h}_{v_{i}}^{(l-1)}, 
 &\mathbf{k}_{ij,c}^{(l)} = \mathbf{W}_{k,c}^{(l)}  \mathbf{z}_{ij}^{(l)}, 
    \mathbf{v}_{ij,c}^{(l)} = \mathbf{W}_{v,c}^{(l)}  \mathbf{z}_{ij}^{(l)}, 
\end{split}
\vspace{-2mm}
\end{equation}
where 
$||$ indicates concatenation operation, $\mathrm{MLP}_{Z}^{(l)}$ is a multi-layer perception network (MLP) with one hidden layer, $\mathbf{W}_{q,c}^{(l)},  \mathbf{W}_{k,c}^{(l)}, \mathbf{W}_{v,c}^{(l)} \in  \mathbb{R}^{\frac{d}{C} \times d}$ are learnable parameters, and $C$ is the number of attention heads.  Specially, we have $\mathbf{h}^{(0)}_{v_{i}} =  \mathbf{W}_{r}\mathbf{x}_{v_{i}}$, where $\mathbf{W}_{r}\in \mathbb{R}^{d \times d_{V_R}}$ are learnable parameters to transform node attribute $\mathbf{x}_{v_{i}}$.

Furthermore, to reduce the computation cost, inspired by \cite{shirzad2023exphormer}, we replace fully-connected graph attention with localized neighborhood attention augmented by a virtual molecule node. 
Specifically, \ring adds a virtual molecule node to $\GR$ and connects the virtual node to all other ring nodes using virtual edges. In this way, global information can be gathered and spread through the virtual node without the need to compute attention between all pairs of nodes.  We initialize the representation of the virtual node using learnable embedding and set the edge attributes of the virtual edges as one-hot vectors different from real edges in $\GR$. Then, we calculate localized neighborhood attention between the target node and its immediate neighbors as follows:  
\begin{equation}\label{eq:attention}
\small
    \alpha_{ij}= \frac{\langle\mathbf{q}_{i,c}^{(l)},\mathbf{k}_{ij,c}^{(l)}\rangle}{\sum_{v_{k}\in \mathcal{N}_{\GR}(v_{i})}\langle\mathbf{q}_{i,c}^{(l)},\mathbf{k}_{ik,c}^{(l)}\rangle},
\end{equation}
where $\mathbf{q}_{i,c}^{(l)}, \mathbf{k}_{ij,c}^{(l)}, \mathbf{v}_{ij,c}^{(l)} \in \mathbb{R}^{\frac{d}{C}}$ are the query, key, value vectors calculated in Eq.\eqref{eq:tokens}, $\mathcal{N}_{\GR}(v_{i})$ is the neighborhood of node $v_{i}$ in $\GR$, and $\langle\mathbf{q},\mathbf{k}\rangle=\frac{{\mathbf{q}}^{T} \mathbf{k}}{\sqrt{d}}$.
Then, we update ring node representations based on cross-attention as follows:
\begin{equation}
\small
\hat{\mathbf{h}}^{(l)}_{v_{i}, R} = \mathbf{W}_{s}^{(l)}\mathbf{h}_{v_{i}}^{(l-1)} 
    +\mathbf{W}_{o}^{(l)} \bigparallel_{c=1}^C(\sum_{v_{j}\in \mathcal{N}_{\GR}(v_{i})}\alpha_{ij}\mathbf{v}_{ij,c}^{(l)}),
\end{equation}
where $\mathbf{W}_{s}^{(l)}, \mathbf{W}_{o}^{(l)} \in \mathbb{R}^{d \times d}$ are learnable parameters,  $\hat{\mathbf{h}}^{(l)}_{v_{i}, R} \in \mathbb{R}^{d}$ is the output node representation of node $v_{i}$ after multi-head cross-attention. Specially, we calculate 
$\mathbf{h}_{v_{i}}^{(0)} = \mathbf{W}_{r}\mathbf{x}_{v_{i}}||\mathbf{p}_{v_{i}}$, where $\mathbf{p}_{v_{i}}\in \mathbb{R}^{d_{p}}$ is a real-valued embedding vector with dimension $d_{p}$ working as node $v_{i}$'s degree-based position encoding~\cite{ying2021transformers} and $\mathbf{W}_{r}\in \mathbb{R}^{(d-d_{p}) \times d_{V_R}}$ is learnable parameter.

Following the convention of typical transformer blocks~\cite{vaswani2017attention}, the node representations generated by cross-attention further go through a feed-forward layer and are updated as follows:
\begin{equation}\label{eq:ffn}
\small
\mathbf{h}^{(l)}_{v_{i}, R} = \mathrm{FFN}^{(l)}(\hat{\mathbf{h}}_{v_{i}, R}^{(l)}+ \mathbf{h}_{v_{i}}^{(l-1)})
\end{equation}
where $\mathrm{FFN}^{(l)}$ is the feed-forward block in the $l$-th layer, and $\mathbf{h}^{(l)}_{v_{i}, R} \in \mathbb{R}^{d}$ is the output node representation of \ring in the $l$-th \algo layer. 

\header\textbf{Inter-level Message Passing Module.} 
In the \inter, we further transfer knowledge between atom and ring nodes on the bipartite graph $\GI$, allowing atom representations to perceive the global structure of the high-level ring system, while ring representations are enriched by the local structure around their constituent atoms.
Specifically, GNNs are applied on the \inter as follows: 
\begin{equation}\label{eq:gnn_i}
\small
    \mathbf{h}^{(l)}_{v_{i},I} = \mathrm{GNN}_{I}^{(l)}(\mathbf{h}_{v_{i}}^{(l-1)}, \{\mathbf{h}_{v_{j}}^{(l-1)}|v_{j}\in \mathcal{N}_{\GI}(v_{i})\}), 
\end{equation}
where $v_{i}$, $v_{j} \in V_{I}$ denotes nodes in $\GI$, 
$\mathcal{N}_{\GI}(v_{i})$ is the neighborhood of node $v_{i}$ in $\GI$, and $\mathbf{h}^{(l)}_{v_{i},I}\in \mathbb{R}^{d}$ is the output node representation of \inter in the $l$-th \algo layer.

\header\textbf{Hierarchical Messages Fusion.} 
After learning node representations on the three levels of graphs, the $l$-th \algo layer generates two types of node representations from different graph hierarchies for every atom and ring node. 
Specially, an atom node $v_{i} \in V_{A}$ has $\mathbf{h}_{v_{i}, A}^{(l)}$ generated by \atom on  $\GA$ and $\mathbf{h}_{v_{i}, I}^{(l)}$ generated by \inter on $\GI$.
Similarly, an ring node $v_{j} \in V_{R}$ has $\mathbf{h}_{v_{j}, R}^{(l)}$ learned by \ring on $\GR$ and $\mathbf{h}_{v_{j}, I}^{(l)}$ learned by \inter on $\GI$.

To facilitate information fusion across the hierarchies, we combine the two representations for each node:

\begin{equation}
\small
    \mathbf{h}^{(l)}_{v_{i}} =  \left\{\begin{array}{lr}
    \mathrm{MLP}_{A}^{(l)}(\mathbf{h}^{(l)}_{v_{i},A} ||\mathbf{h}^{(l)}_{v_{i},I} )  & \text{if $v_{i}\in V_{A}$} \\
    \\
    \mathrm{MLP}_{R}^{(l)}(\mathbf{h}^{(l)}_{v_{i}, R} ||\mathbf{h}^{(l)}_{v_{i},I} ) & \text{if $v_{i}\in V_{R}$}\\
    \end{array}\right.
\end{equation}
where  $\mathbf{h}^{(l)}_{v_{i}}\in \mathbb{R}^{d} $ is the final node representation output by the $l$-th \algo layer for node $v_{i} \in V_{A} \cup V_{R}$.  $\mathbf{h}^{(l)}_{v_{i}}$ is then sent to the next \algo layer for further update.

\subsection{Prediction Layer}\label{sec::prediction}
After stacking $L$ \algo layers, we aggregate their output node representations to generate graph representation for the \graph. First, for each node $v_{i} \in V_{A} \cup V_{R}$, we concatenate its representations from all \algo layers as its final node representation to incorporate structural patterns at different scales:
\begin{equation}
\small
    \mathbf{h}_{v_{i}} = \mathrm{CONCAT}(\mathbf{h}_{v_{i}}^{(0)}, \mathbf{h}_{v_{i}}^{(1)}, \cdots, \mathbf{h}_{v_{i}}^{(L)}),
\end{equation}
where $\mathrm{CONCAT}$ indicates concatenation operation, and $\mathbf{h}_{v_{i}} \in \mathbb{R}^{d \times (L+1)}$ is the concatenated node representation. Then, we separately aggregate node representations of atoms and rings in $\G$ using sum pooling and concatenate the two to obtain the final graph representation:
\begin{equation}
\small
\mathbf{h}_{\mathcal{G}} =  \mathrm{POOL}(\{\mathbf{h}_{v_{i}}|v_{i} \in V_{A}\}) || \mathrm{POOL}(\{\mathbf{h}_{v_{i}}|v_{i} \in V_{R}\}),
\end{equation}
where $\mathrm{POOL}$ indicates sum pooling operation and $\mathbf{h}_{\mathcal{G}} \in \mathbb{R}^{2d \times (L+1)}$ is the final graph representation for the \graph $\G$. As the target properties of OSC molecules considered in this paper are real values, and thus OSC property prediction can be regarded as a regression task, we project the molecular representation into the logits $\hat{y}_{\mathcal{G}}$ using one linear layer. Given a batch of training molecules,  \algo is trained with mean absolute error (MAE) loss: 
\begin{equation}\label{eq:loss}
\small
\mathcal{L} = \frac{1}{B}\sum_{b=1}^{B}|y_{\mathcal{G}_{b}} - \hat{y}_{\mathcal{G}_{b}}|,
\end{equation}
where $B$ is the batch size, and $\hat{y}_{\mathcal{G}{b}}$ and $y_{\mathcal{G}{b}}$ represent the predicted and ground-truth molecule properties, respectively.

%% file: experiments.tex
\section{Experiments}\label{sec:experiments}
\begin{table}[!tp]
    \centering
    \caption{Dataset statistics.}
    \label{tab:dataset}
    \vspace{-2mm}
    \setlength{\tabcolsep}{2pt}
\def\arraystretch{0.92}
\resizebox{1.0\linewidth}{!}{
    \begin{tabular}{ccccc}
    \toprule
    DATASET & \# GRAPHS & AVG. \# NODES & AVG. \# EDGES & AVG. \# RINGS \\
    \midrule
    \cepdb & 2.3M  & 27.6  & 33.3  & 6.7  \\
    \hopv & 350   & 42.7  & 49.3  & 7.5  \\
    \polymer & 1055  & 77.1  & 84.2  & 8.2  \\
    \nnfa & 654   & 118.2  & 133.0  & 15.8  \\
    \pnfa & 277   & 80.7  & 88.2  & 8.5  \\
    \bottomrule
    \end{tabular}
    }
\vspace{-4mm}
\end{table}
\subsection{Experimental Settings}\label{sec::datasets}
\paragraph{Datasets and Evaluation Metrics.}
 We curate 5 OSC molecule datasets to evaluate property prediction performance, as listed in Table~\ref{tab:dataset}. Specially, \cepdb~\cite{hachmann2011harvard} dataset~
is generated based on density functional theory (DFT),  and other four datasets, \hopv~\cite{lopez2016harvard}, \polymer~\cite{nagasawa2018computer}, \nnfa~\cite{miyake2021machine}, and \pnfa~\cite{miyake2021machine}, are datasets consisting of different types of OSC molecules whose properties are experimentally validated. Additional dataset and target OSC properties descriptions can be found in Appendix. We adopted scaffold-based splitting~\cite{wu2018moleculenet}, commonly used in the field, to partition molecules into training, validation, and testing sets with a ratio of 6:2:2. We use mean absolute error (MAE) as the evaluation metric following previous studies~\cite{eibeck2021predicting,saleh2023energy}.

\paragraph{Baselines and Implementation Details.}
We compare  \algo with 11 competitors in 4 categories. (i) Fingerprint-based methods, \emph{\maccs}~\cite{durant2002reoptimization}, and \emph{\ecfp}~\cite{rogers2010extended}.
(ii) GNN-based methods,   \emph{\gine}~\cite{hu2019strategies}, \emph{\gvn}~\cite{gilmer2017neural}, \emph{\afp}~\cite{xiong2019pushing}, and \emph{\ognn}~\cite{zhu2023o}. (iii) Pooling-based methods, including \emph{\topk}~\cite{gao2019graph} and \emph{\sag}~\cite{lee2019self}. 
(iv) Transformer-based methods,    \emph{\graphormer}~\cite{ying2021transformers}, \emph{\gps}~\cite{rampavsek2022recipe}, and  \emph{\vit}~\cite{he2023generalization}. 
The implementation details of \algo and other baselines are provided in Appendix.

\begin{table}[!tp]
  \centering
  \caption{PCE (\%)  prediction performance compared between \algo and  baselines in terms of test MAE (↓). 
  ↓ indicates smaller values are better. 
The mean and standard deviation are reported. 
\textbf{Bold}: best. \underline{Underline}: runner-up.
}
\vspace{-2mm}
  \setlength{\tabcolsep}{2.5pt}
\def\arraystretch{0.92}
\resizebox{1.0\linewidth}{!}{
\begin{tabular}{cccccc}
    \toprule
          Method & \cepdb & \hopv & \polymer & \nnfa & \pnfa \\
    \midrule
    \maccs & 0.898\scriptsize{±0.001} & 1.632\scriptsize{±0.008} & \textbf{1.770\scriptsize{±0.016}} & 2.614\scriptsize{±0.015} & 2.594\scriptsize{±0.023} \\
    \ecfp & 0.510\scriptsize{±0.001} & 1.544\scriptsize{±0.026} & 1.787\scriptsize{±0.012} & \underline{2.377\scriptsize{±0.024}} & 2.704\scriptsize{±0.016} \\
    \midrule
    \gine & 0.460\scriptsize{±0.006} & 1.614\scriptsize{±0.021} & 1.826\scriptsize{±0.013} & 2.620\scriptsize{±0.035} & 2.528\scriptsize{±0.064} \\
    \gvn  & 0.393\scriptsize{±0.011} & 1.724\scriptsize{±0.020} & 1.878\scriptsize{±0.017} & 3.164\scriptsize{±0.174} & 2.962\scriptsize{±0.207} \\
    \afp  & 0.415\scriptsize{±0.009} & 2.002\scriptsize{±0.032} & 1.897\scriptsize{±0.099} & 2.826\scriptsize{±0.082} & 2.608\scriptsize{±0.029} \\
    \ognn & 0.267\scriptsize{±0.004} & 1.727\scriptsize{±0.073} & 1.868\scriptsize{±0.063} & 2.587\scriptsize{±0.179} & 2.866\scriptsize{±0.288} \\
    \midrule
    \topk & 0.527\scriptsize{±0.046} & 1.598\scriptsize{±0.013} & 1.830\scriptsize{±0.008} & 2.644\scriptsize{±0.045} & 2.523\scriptsize{±0.064} \\
    \sag  & 0.536\scriptsize{±0.044} & 1.607\scriptsize{±0.057} & 1.841\scriptsize{±0.020} & 2.648\scriptsize{±0.101} & 2.557\scriptsize{±0.057} \\
    \midrule
    \gps  & 0.247\scriptsize{±0.010} & 1.942\scriptsize{±0.128} & 2.395\scriptsize{±0.221} & 3.233\scriptsize{±0.208} & 2.690\scriptsize{±0.186} \\
    \graphormer & -     & 1.609\scriptsize{±0.069} & 1.799\scriptsize{±0.047} & 2.689\scriptsize{±0.096} & \underline{2.522\scriptsize{±0.058}} \\
    \vit  & \underline{0.244\scriptsize{±0.009}} & \underline{1.479\scriptsize{±0.061}} & 1.887\scriptsize{±0.083} & 2.467\scriptsize{±0.166} & 2.856\scriptsize{±0.089} \\
    \midrule
    \algo & \textbf{0.189\scriptsize{±0.003}} & \textbf{1.477\scriptsize{±0.021}} & \underline{1.776\scriptsize{±0.014}} & \textbf{2.259\scriptsize{±0.012}} & \textbf{2.482\scriptsize{±0.047}} \\
    \bottomrule
    \end{tabular}
    }
  \label{tab:PCE}
    \vspace{-6mm}
\end{table}

\subsection{Overall Performance}
\noindent \textbf{PCE prediction.} We report the performance of \algo in predicting power conversion efficiency (PCE), the primary property of interest for OSC molecules~\cite{solakAdvancesOrganicPhotovoltaic2023}, in Table~\ref{tab:PCE}. Firstly, we observe that \algo consistently achieves the best performance, except that \algo is the runner-up in \polymer. For instance, on \cepdb, \algo achieves test MAE $0.189$, which indicates $22.8\%$ relative improvement over the best competitor with test MAE $0.244$. On \nnfa, the dataset with the highest average number of rings (See Table~\ref{tab:dataset}), \algo outperforms the fingerprint-based method \ecfp by $4.96\%$. In contrast, other deep learning models struggle to match the performance of \ecfp.
Furthermore, across the four experimental datasets (\hopv, \polymer, \nnfa, \pnfa), 
we observe that GNN-based methods consistently yield inferior performance compared to fingerprint-based methods, indicating a struggle in learning structural patterns from larger and more complicated OSC molecules.
However, \algo still achieves competitive performance across these datasets, emerging as the top performer in three out of four datasets. The results demonstrate \algo's ability to capture structural patterns in OSC molecules for accurate property prediction.

\begin{table}[!t]
  \centering
  \caption{Multi-task learning performance on \cepdb by test MAE (↓). The second row indicates the units of different properties. \textbf{Bold}: best. \underline{Underline}: runner-up.}
  \vspace{-2mm}
  \setlength{\tabcolsep}{1pt}
\def\arraystretch{0.92}
  \resizebox{1.0\linewidth}{!}{
        \begin{tabular}{ccccccc}
    \toprule
    \multirow{2}[4]{*}{Method} & PCE   & HOMO  & LUMO  & Band Gap & Voc   & Jsc \\
\cmidrule{2-7}          & $\%$  & $eV$  & $eV$  & $eV$  & $V$   & $mA/cm^2$ \\
    \midrule
    \maccs & 0.898\scriptsize{±0.001} & 0.115\scriptsize{±0.001} & 0.122\scriptsize{±0.001} & 0.184\scriptsize{±0.001} & 0.115\scriptsize{±0.001} & 35.297\scriptsize{±0.002} \\
    \ecfp & 0.510\scriptsize{±0.001} & 0.066\scriptsize{±0.001} & 0.065\scriptsize{±0.001} & 0.080\scriptsize{±0.001} & 0.066\scriptsize{±0.001} & 15.574\scriptsize{±0.003} \\
    \midrule
    \gine & 0.491\scriptsize{±0.007} & 0.049\scriptsize{±0.001} & 0.058\scriptsize{±0.001} & 0.073\scriptsize{±0.001} & 0.049\scriptsize{±0.001} & 15.409\scriptsize{±0.292} \\
    \gvn  & 0.496\scriptsize{±0.007} & 0.048\scriptsize{±0.001} & 0.059\scriptsize{±0.003} & 0.073\scriptsize{±0.003} & 0.048\scriptsize{±0.001} & 15.110\scriptsize{±0.141} \\
    \afp  & 0.453\scriptsize{±0.018} & 0.041\scriptsize{±0.003} & 0.057\scriptsize{±0.003} & 0.068\scriptsize{±0.004} & 0.040\scriptsize{±0.003} & 14.182\scriptsize{±0.548} \\
    \ognn & 0.259\scriptsize{±0.008} & 0.026\scriptsize{±0.001} & 0.030\scriptsize{±0.001} & 0.036\scriptsize{±0.001} & 0.026\scriptsize{±0.001} & 8.039\scriptsize{±0.004} \\
    \midrule
    \topk & 0.566\scriptsize{±0.047} & 0.053\scriptsize{±0.003} & 0.060\scriptsize{±0.009} & 0.078\scriptsize{±0.007} & 0.053\scriptsize{±0.003} & 16.120\scriptsize{±0.859} \\
    \sag  & 0.581\scriptsize{±0.009} & 0.056\scriptsize{±0.001} & 0.050\scriptsize{±0.001} & 0.074\scriptsize{±0.001} & 0.056\scriptsize{±0.001} & 15.510\scriptsize{±0.034} \\
    \midrule
    \gps  & \underline{0.241\scriptsize{±0.018}} & \underline{0.020\scriptsize{±0.003}} & \underline{0.021\scriptsize{±0.001}} & \underline{0.025\scriptsize{±0.001}} & \underline{0.018\scriptsize{±0.002}} & \underline{7.514\scriptsize{±0.350}} \\
    \vit  & 0.322\scriptsize{±0.050} & 0.040\scriptsize{±0.007} & 0.035\scriptsize{±0.009} & 0.055\scriptsize{±0.007} & 0.040\scriptsize{±0.007} & 12.479\scriptsize{±1.939} \\
    \midrule
    \algo & \textbf{0.193\scriptsize{±0.007}} & \textbf{0.014\scriptsize{±0.001}} & \textbf{0.018\scriptsize{±0.001}} & \textbf{0.023\scriptsize{±0.001}} & \textbf{0.014\scriptsize{±0.001}} & \textbf{5.993\scriptsize{±0.304}} \\
    \bottomrule
    \end{tabular}
    }
  \label{tab:multitask}
    \vspace{-6mm}
\end{table}

\header\textbf{Multi-task learning.}
We further evaluate the performance of \algo in multi-task learning using \cepdb dataset. Specifically, we aim to predict 5 target properties, resulting in 5 regression tasks. The details of the five target properties are given in Appendix. For training \algo and other deep neural network-based competitors, 
we set the output dimension to be the same as the number of target properties and train the neural network using MAE loss. As the results shown in Table~\ref{tab:multitask}, \algo consistently outperforms other competitors in all six target properties,  often by a significant margin.  For instance, \algo has test MAE $5.993$ in predicting \jsc, achieving 20.24\% relative improvement on the best competitor. The results further validate \algo's superiority in predicting multiple OSC properties simultaneously. 
Additionally, we observe \gps consistently achieve promising results across all target properties, only inferior to \algo. It is because both \algo and \gps combine the power of message passing and global attention, which validates the importance of capturing both local and global structural features in OSC molecules.
\subsection{Model Analysis} 

\noindent \textbf{Ablation on \graph.} We evaluate the effectiveness of  \graph by comparing  it with all three levels $\mathcal{G}=\{\GA, \GR, \GI\}$ to a subset of $\mathcal{G}$. As shown in Table~\ref{tab:subgraph},  we observe a significant performance drop on the incomplete \graph compared to the full one, which validates the necessity of all levels in $\mathcal{G}$. Moreover, observing that $\mathcal{G} \backslash \GI$ is better than  $\GA$ and $\GR$, we conclude that encoding structure of both atom and ring level is important in 
OSC property prediction.
Comparing $\G$ with $\G \backslash \GI$, we find transferring hierarchical information between atom and ring levels can   improve   performance.

\begin{table}[]
    \centering
    \caption{Ablation on \graph by test MAE (↓). $ \backslash$ denotes exclusion. }
    \vspace{-3mm}
      \setlength{\tabcolsep}{2.5pt}
\def\arraystretch{0.92}
    \resizebox{0.62\linewidth}{!}{
    \begin{tabular}{cccccc}
    \toprule
          & \cepdb & \hopv & \polymer & \nnfa & \pnfa \\
    \midrule
     $\GA$ only & 0.550  & 1.912  & 1.841 & 3.122  & 2.683  \\
    $\GR$ only & 0.358  & 1.874  & 2.001  & 2.412  & 2.531  \\
    $\G \backslash \GR$ & 0.606  & 1.526  & 1.831  & 2.304  & 2.540  \\
    $\G \backslash \GI$ & \underline{0.315 } & \underline{1.497 } & \underline{1.795}  & \underline{2.299 } & \underline{2.501 } \\
    \midrule
    $\G$  & \textbf{0.189 } & \textbf{1.477 } & \textbf{1.776 } & \textbf{2.259 } & \textbf{2.482 } \\
    \bottomrule
    \end{tabular}
    }
  \label{tab:subgraph}
  \vspace{-2mm}
\end{table}

\begin{table}[!tp]
    \centering

  \caption{PCE prediction performance of different implementations of ring-level graph encoder by test MAE (↓).}
  \vspace{-2mm}
  \setlength{\tabcolsep}{1mm}
  \resizebox{\linewidth}{!}{
    \begin{tabular}{cccccc}
    \toprule
      Encoder    & \cepdb & \hopv & \polymer & \nnfa & \pnfa \\
    \midrule
    Cross-attention & \underline{0.1886 } & \textbf{1.4774 } & \textbf{1.7757 } & \underline{2.2588 } & \textbf{2.4819 } \\
    Cross-attention w.o. virtual & \textbf{0.1860 } & 1.5106  & 1.8082  & 2.2602  & 2.5876  \\
    \midrule
    \gine & 0.3576  & \underline{1.4796 } & 1.7892  & \textbf{2.2519 } & 2.6203  \\
    \gvn  & 0.3136  & 1.5069  & \underline{1.7856 } & 2.2665  & \underline{2.5296 } \\
    \transformer & 0.2319  & 1.5394  & 1.8273  & 2.4151  & 2.5805  \\
    \gps  & 0.2231  & 1.5137  & 1.7910  & 2.4069  & 2.5390  \\
    \bottomrule
    \end{tabular}
    }
  \label{tab:ring}
  \vspace{-2mm}
\end{table}
\begin{table}[!t]
    \centering

  \caption{Comparison of training time per epoch and inference time per epoch of different implementations of ring-level graph encoder by seconds (s).}
  \vspace{-2mm}
  \setlength{\tabcolsep}{1pt}
  \resizebox{\linewidth}{!}{
    \begin{tabular}{ccccccccccc}
    \toprule
          & \multicolumn{2}{c}{\cepdb} & \multicolumn{2}{c}{\hopv} & \multicolumn{2}{c}{\polymer} & \multicolumn{2}{c}{\nnfa} & \multicolumn{2}{c}{\pnfa} \\
\cmidrule{2-11}          & Test  & Train & Test  & Train & Test  & Train & Test  & Train & Test  & Train \\
    \midrule
    \algo & 71.27  & 504.1  & 0.037  & 0.671  & 0.097  & 1.349  & 0.103  & 0.933  & 0.033  & 0.407  \\
    \algo w.o. virtual & 65.59  & 459.9  & 0.032  & 0.434  & 0.093  & 1.288  & 0.099  & 0.883  & 0.032  & 0.386  \\
    \midrule
    \gine & 63.34 & 444.5 & 0.030  & 0.367  & 0.089  & 1.058  & 0.093  & 0.782  & 0.030  & 0.323  \\
    \gvn  & 65.24 & 440.2 & 0.031  & 0.480  & 0.091  & 1.095  & 0.095  & 0.792  & 0.031  & 0.332  \\
    \transformer & 103.5 & 764.5 & 0.048  & 0.639  & 0.109  & 1.704  & 0.116  & 1.179  & 0.042  & 0.507  \\
    \gps  & 85.71  & 660.1 & 0.048  & 0.687  & 0.118  & 1.948  & 0.123  & 1.353  & 0.047  & 0.599  \\
    \bottomrule
    \end{tabular}
    }
  \label{tab:ring_efficiency}
  \vspace{-3mm}
\end{table}

\header\textbf{Effectiveness of \ring.}
We evaluate the effectiveness of the proposed \ring by replacing it with other graph representation learning layers and report the results in Table~\ref{tab:ring}. Specifically, \textit{Cross-attention w.o. virtual} is the variant of our cross-attention without virtual molecule nodes. \textit{Vanilla Transformer}~\cite{vaswani2017attention} is the transformer encoder without special designs for graphs. The results in 
Table~\ref{tab:ring} shows that our proposed cross-attention achieves the best performance in 3 out of 5 datasets and competitive results in the other 2 datasets, which validates its effectiveness. We further compare the training time per epoch and inference time per epoch in seconds
with other graph representation learning layers, with results in Table~\ref{tab:ring_efficiency}. We observe that \ring requires significantly less time than transformer-based methods for training and inference, indicating that \ring can reduce computational costs in transformers by replacing full-connected attention with local neighborhood attention.

\begin{table}[!tp]
\centering
     \caption{Comparison between using rings and BRICS-based motifs as high-level structure by test MAE (↓). }\label{tab:motif}

 \vspace{-2mm}
 \setlength{\tabcolsep}{2pt}
\def\arraystretch{0.95}
\resizebox{0.8\linewidth}{!}{
    \begin{tabular}{lccccc}
    \toprule
          & CEPDB & HOPV  & PolymerFA & nNFA  & pNFA \\
    \midrule

    Motifs &  0.4556
     & 1.5203  & \textbf{1.7230}  & 2.5013  & 2.5960  \\
    Rings & \textbf{0.1886}  & \textbf{1.4774}  & 1.7757  & \textbf{2.2588}  & \textbf{2.4819}  \\
    Ring+Motifs & 0.2440 & 1.4833  & 1.8098  & 2.7561  & 2.5908  \\
    \bottomrule
    \end{tabular}
    }
\vspace{-4mm}
\end{table}

\header\textbf{Comparison to molecular motifs.}
As rings can be considered as a special kind of chemical motif, one question is how our method will perform when using  general chemical motifs instead of focusing on rings. Hence, we explore the effect of modeling general chemical motifs. Following \citet{zhang2021motif}, we extract molecular motifs using BRICS~\cite{degen2008art}, 
and replace the \gr in \algo with a motif-level graph where motifs are regarded as nodes. 

We further combine BRICS functional group motifs with ring systems in the high-level graph (Ring+Motif), connecting functional group nodes with ring nodes if they share atoms.
As results shown in Figure~\ref{tab:motif}, we observe that \algo generally achieves better performance by modeling rings than motifs  and combination of rings and motifs.
The results indicate the importance of focusing on the patterns in ring systems for OSC property prediction.

\begin{figure}[!tp] 
	\centering
		\includegraphics[width=0.82\linewidth]{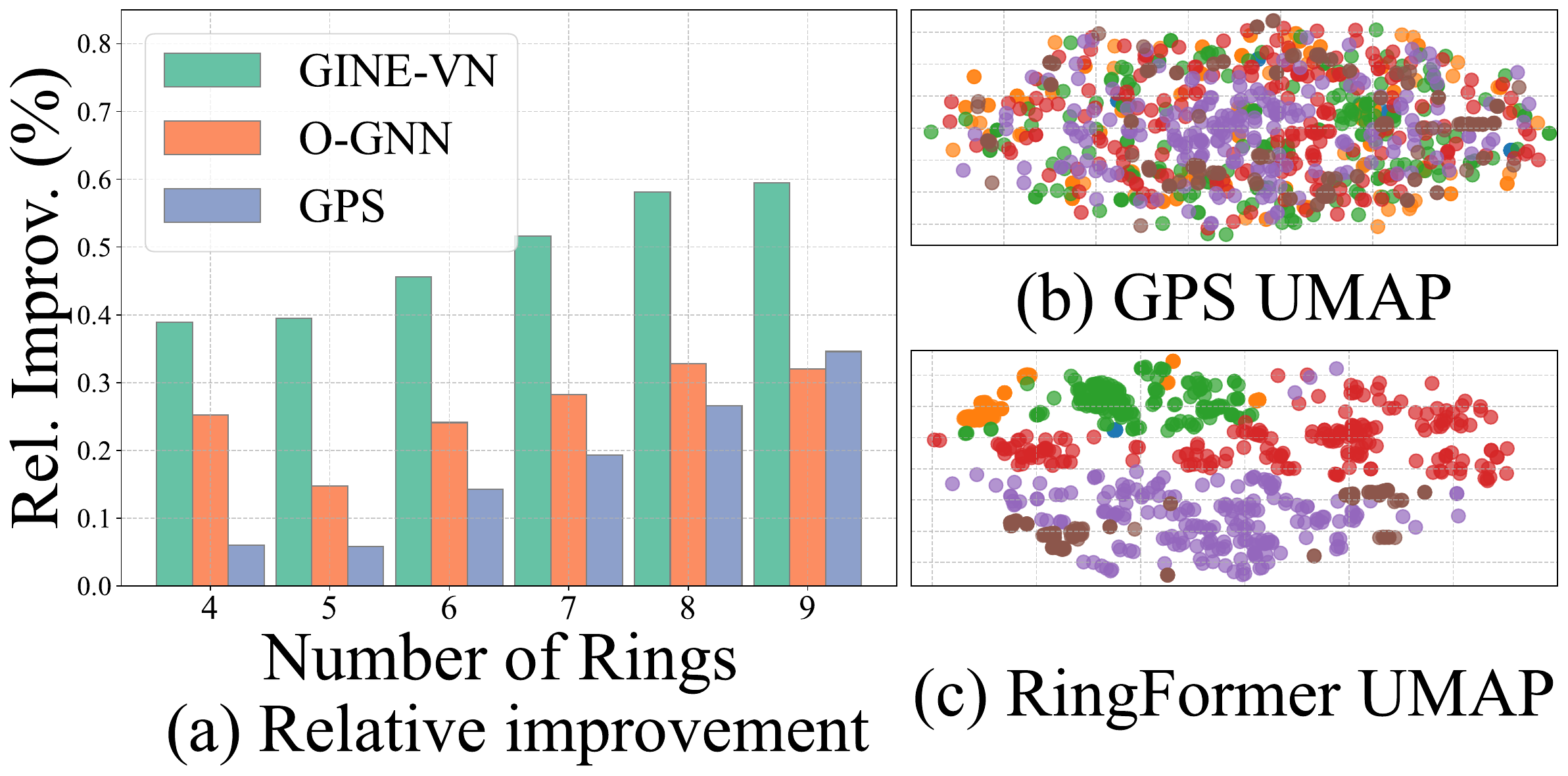}

  \vspace{-2mm}
\caption{(a) Performance improvement on molecules with varying numbers of rings; (b) and (c) Comparison of embedding visualizations, colors representing the number of rings.}
\label{fig:umap}
\vspace{-6mm}
\end{figure}

\header\textbf{Performance \textit{vs.} ring system complexity.}
To demonstrate \algo's advantage in leveraging ring structures within OSC molecules,  we evaluate the performance of \algo on OSC molecules with different numbers of rings, which reflect different levels of ring system complexity.
Firstly, we examine the relative improvement in MAE achieved by \algo in predicting PCE for molecules with varying numbers of rings in the \cepdb test set, compared to baseline methods \gvn, \ognn, and \gps. As illustrated in Figure~\ref{fig:umap} (a), the performance gain generally increases together with the number of rings in molecules, indicating a clear correlation between \algo's superiority and the complexity of the ring systems.
Furthermore, we visualize the graph representations of OSC molecules in the \cepdb test set using UMAP~\cite{mcinnes2018umap}.
As depicted in Figure~\ref{fig:umap} (b) and (c), the embeddings generated by \algo can be distinctly separated based on the number of rings in the OSC molecules, unlike those generated by \gps. 
These observations confirm that \algo excels in capturing the intricate structures of ring systems.

\header 
\textbf{More experiments.} In  Appendix, we present the performance of \algo with different GNN backbones other than \gine in Table \ref{tab:backbone}, and we report the results of varying the number of \algo layers in Figure~\ref{fig:layer}.

%% file: conclusion.tex
\section{Conclusion}

This paper addresses the under-explored problem of predicting properties of organic solar cells (OSCs) and introduces \algo, a novel graph transformer framework designed to capture rings and their interconnections within an OSC molecule to facilitate accurate prediction. \algo constructs a \graph that represents OSC molecular structure at both atom and ring levels, and leverages a combination of local message-passing and global attention mechanisms to learn expressive graph representations. 
Extensive experiments demonstrate the superiority of \algo  in OSC property prediction.

%% file: appendix.tex
\clearpage  

\appendix  
\section*{Appendix}

\section{Dataset Details}\label{apx:dataset}
\header \textbf{\cepdb}: This dataset is curated from the Harvard's clean energy project database~\cite{hachmann2011harvard} which contains approximately 2.3 million organic semiconductor molecules with potential applications as donor materials in organic solar cells. Each molecule is initially represented by a SMILES string, and the database includes OSC-related properties calculated using density functional theory (DFT) methods. During dataset curation, we eliminate invalid SMILES strings using RDKit\footnote{\url{https://www.rdkit.org/docs/index.html}} and resulting in a final set of 2,225,974 valid molecules. Six properties are included for OSC property prediction: power conversion efficiency (PCE), highest occupied molecular orbital (HOMO), lowest unoccupied molecular orbital (LUMO), Band Gap, open circuit voltage (\voc), and short-circuit current density (\jsc). The SMILES strings are converted to atom-level molecular graphs using PyTorch-Geometric\footnote{\url{https://pytorch-geometric.readthedocs.io/en/latest/index.html}}.

\header \textbf{\hopv}: The Harvard Organic Photovoltaic (\hopv) dataset~\cite{lopez2016harvard} collects the data of 350 selected small molecules and polymers, represented by SMILES strings, intended for use as p-type constituents in OSCs. The dataset encompasses various experimentally determined properties. During dataset curation, PCE is regarded as the ground truth property, and the SMILES strings are converted to atom-level molecular graphs.

\header \textbf{\polymer}: This dataset is originally reported by Nagasawa \etal \cite{nagasawa2018computer}, 
comprises 1203 donor polymer molecules with various device parameters, including PCE. These device parameters are experimentally obtained from old-generation polymer:fullerene-based OSC devices. Each molecule is assigned a SMILES string and a nickname. Due to potential duplications arising from one donor polymer molecule working with different acceptors, resulting in different PCE values, molecules with the same nickname are merged. The largest PCE value is considered the ground truth, and after SMILES validity checks, 1055 molecules remain, represented by molecular graphs.

\header \textbf{\nnfa}:  This dataset is curated from the Polymer:NFA dataset reported by Miyake \etal~\cite{miyake2021machine} containing 1318 pairs of polymer donor and non-fullerene acceptor (NFA) molecules, along with experimentally determined PCE values. NFAs represent a new generation of electron acceptors for organic photovoltaics, distinct from fullerene structures, offering significantly enhanced performance for OSC devices. During dataset curation, pairs of molecules with the same acceptor molecule are merged based on SMILES strings, and the largest PCE value is considered the ground truth. After SMILES validity checks, 654 acceptor molecules remain, represented by molecular graphs.

\header \textbf{\pnfa}:  \pnfa dataset, also obtained from the Polymer:NFA dataset~\cite{miyake2021machine}, differs from \nnfa in that it focuses on predicting OSC properties based on polymer donor molecules. During dataset curation, pairs of molecules with the same donor molecule are merged based on SMILES strings, and the largest PCE value is considered the ground truth. After SMILES validity checks, 277 donor molecules remain, represented by molecular graphs.

\section{OSC Properties Details}
In this paper, we focus on six key OSC properties: power conversion efficiency (PCE), highest occupied molecular orbital (HOMO), lowest unoccupied molecular orbital (LUMO), band gap, open-circuit voltage (\voc), and short-circuit current density (\jsc). Below, we provide detailed descriptions of each property:
\begin{itemize}
    \item \textbf{PCE} ($\%$) of an organic solar cell is expressed as the percentage ratio of electrical power produced to optical power impinging on the cell~\cite{sikiru2022recent}. It is the most critical component of every OSC system, ranging from 0\% to 100\%.
    \item The \textbf{HOMO} ($eV$) is the highest-energy molecular orbital that has electrons in it and the \textbf{LUMO} ($eV$) is the next energy orbital level close to HOMO, which always has states that are empty of electrons~\cite{hussain2018handbook}.
    \item \textbf{Band Gap} ($eV$) or HOMO–LUMO gap is the energy difference between the HOMO and LUMO. Its size can be used to predict the strength and stability of transition metal complexes~\cite{griffith1957ligand}. As a rule of thumb, the smaller a compound's HOMO–LUMO gap, the more stable the compound.
    \item $\mathbf{V_{oc}}$ ($V$) is the maximum voltage available from a solar cell, and this occurs at zero current~\cite{sebestyen2021renewable}. The open-circuit voltage corresponds to the amount of forward bias on the solar cell due to the bias of the solar cell junction with the light-generated current. 
    \item $\mathbf{J_{SC}}$ ($mA/cm^2$) is the current density at zero voltage~\cite{akhtaruzzaman2021comprehensive}.
\end{itemize}

\section{Implementation Details of Baselines}\label{apx:baselines}
We provide more description and implementation details of the baselines in experiments section.
\begin{itemize}
    \item \textbf{\maccs} is a fingerprint-based method.  \maccs uses fingerprints Molecular ACCess System keys~\cite{durant2002reoptimization} which are one of the most commonly used structural keys. These fingerprints are binary in nature, consisting of a fixed-length bitstring (typically 166 bits). Each bit in the MACCS keys represents the presence or absence of a specific substructure or chemical pattern within a molecule. 
    \item \textbf{\ecfp} is also a fingerprint-based method that uses Extended Connectivity Circular Fingerprints~\cite{rogers2010extended}. ECFP captures the structural information and connectivity patterns of molecules by considering circular neighborhoods around each atom in a molecule. Within these neighborhoods, it identifies and encodes substructures or fragments. The resulting ECFP fingerprint is a binary representation, with each element indicating the presence or absence of specific substructures within the circular neighborhoods. 
    \item \textbf{\gine} indicates graph isomorphism network (GIN)~\cite{xu2018powerful} with edges. It extends the expressive message-passing GNN framework GIN by taking edge attributes in the message passing and achieves promising results in molecule property prediction.
    \item \textbf{\gvn}~\cite{hu2019strategies} is the extension of  \gine that adds a virtual node to the given molecule graph that is connected to all other nodes~\cite{gilmer2017neural}. The virtual node works as a shared global workspace, which every node accesses for both reading and writing during each step of the message passing process.
    \item \textbf{\afp}~\cite{xiong2019pushing} is message-passing GNN specially designed for learning molecular representations. It incorporates graph attention mechanism and virtual node to capture molecular structures. \item \textbf{\ognn}~\cite{zhu2023o} is message-passing GNN that specially considers rings in chemical compounds. Apart from updating node and edge representations in each GNN layer, \ognn specially identifies rings in a molecule and maintains ring representations for them. The ring representations take part in message passing and are updated together with node and edge representations.
    \item \textbf{\topk}~\cite{gao2019graph} is a graph pooling method that adaptively selects nodes to form
    a condensed graph that preserves graph hierarchy. The nodes are selected based on scores predicted by a trainable projector.
    \item \textbf{\sag}~\cite{lee2019self} is a graph pooling method based on self-attention. Its self-attention pools nodes into a smaller graph with graph convolution that takes both node
    features and graph topology into consideration.

    \item \textbf{\graphormer}~\cite{ying2021transformers} is among the first graph transformers that incorporate Transformer architecture in graph representation learning. It designs position encoding for capturing nodes' absolute positions in a graph, and develops spatial encoding to incorporate the relationship between any pair of nodes in the self-attention mechanism.
    \item \textbf{\gps}~\cite{rampavsek2022recipe} is the SOTA graph transformer. It is a graph transformer framework that combines positional/structural encoding,  message-passing mechanism, and global attention mechanism. Each \gps layer contains one GNN layer and one vanilla transformer block working in parallel. 
    \item \textbf{\vit}~\cite{he2023generalization} is a graph transformer that adapts transformer architecture introduced in computer vision to the graph domain. It first fragments the graph into patches and uses GNNs on the patches for learning patch representation, which are then fed to a vanilla transformer for the final graph representation. It shares the benefits of capturing long-range dependencies in graphs.
\end{itemize}

\paragraph{Implementation Details.}
In \algo, we use \gine~\cite{hu2019strategies} as GNNs in \atom and \inter. We fix the number of layers $L=8$, dimension $d=512$, and the number of attention heads $C=4$ in all 5 datasets. 
We use mini-batch gradient descent to optimize parameters in \algo with Adam optimizer and one-cycle learning rate scheduler~\cite{smith2019super} with 5\% percentage of the cycle spent increasing the learning rate. We set the training epochs as 30 and batch size as 1024 in \cepdb. We set the training epochs as 100, and batch size as 32 in the other four datasets.
Maximum learning rate is tuned in $\{0.001, 0.0005, 
0.0001, 0.00005\}$ based on validation set.
For each dataset, we repeat experiments 5 times with different random seeds and report the mean metrics ±  standard deviation.
All experiments are conducted on a Linux server with Intel Xeon Gold 6226R 2.90GHz CPU and an Nvidia RTX 3090 GPU card. 

For fingerprint-based methods \maccs and \ecfp, we first generate fingerprints for molecules according to their principles using DeepChem~\cite{ramsundar2018molecular} and then use the fingerprints to train a random forest regressor, whose effectiveness on OSC property prediction has been widely recognized~\cite{eibeck2021predicting,miyake2021machine}, for predicting molecular properties.  The random forest regressor is implemented using Scikit-learn~\cite{kramer2016scikit}. For deep learning-based methods, including GNN-based methods and transformer-based methods, we implement them based on their official codes following the same training setting as \algo, including batch size, training epochs, optimizer, learning rate scheduler, maximum learning rate, number of trials, and random seeds. For models including \gine, \gvn, \afp, \ognn, \topk, \sag, \graphormer, and \gps, we fix the number of layers as 8,  hidden dimension as 512, and the number of attention heads as 4, which is the same as \algo. For pooling-based methods \topk and \sag, we use \gine as the GNN backbone and pool nodes after the first 4 layers' GNNs with a ratio of 0.5, followed by another 4 layers of GNNs. Other hyperparameters are chosen as suggested in their paper or official codes. For \gps, we use \gine as the GNN backbone. For \vit, we use \gine as the GNN backbone and set the number of GNN layers as 4, the number of transformer layers as 4, the hidden dimension as 512, and the number of attention heads as 4. Other hyperparameters are chosen as suggested in their respective papers or official codes.

\section{Additional Experiments}

\begin{table}[!t]
  \centering
  \caption{Performance of \algo with different GNN backbone by test MAE (↓) in PCE prediction.}
\vspace{-2mm}
 \setlength{\tabcolsep}{4pt}
\def\arraystretch{0.92}
\resizebox{1.0\linewidth}{!}{
    \begin{tabular}{cccccc}
    \toprule
          & \cepdb & \hopv  & \polymer & \nnfa  & \pnfa \\
    \midrule
    Best Deep Competitor & 0.2442  & 1.4792  & 1.7987  & 2.4668  & 2.5221  \\
    \midrule
    GINE  & 0.1886  & \textbf{1.4774 } & \textbf{1.7757 } & \textbf{2.2588 } & 2.4819  \\
    GatedGCN & 0.1899  & 1.5158  & 1.7967  & 2.2742  & \textbf{2.4800 } \\
    GraphSAGE & \textbf{0.1874 } & 1.5393  & 1.7923  & 2.3694  & 2.4982  \\
    \bottomrule
    \end{tabular}}
  \label{tab:backbone}
  \vspace{-2mm}
\end{table}

 \begin{figure*}[!tp] 
    \centering 
    \includegraphics[width=0.95\linewidth]{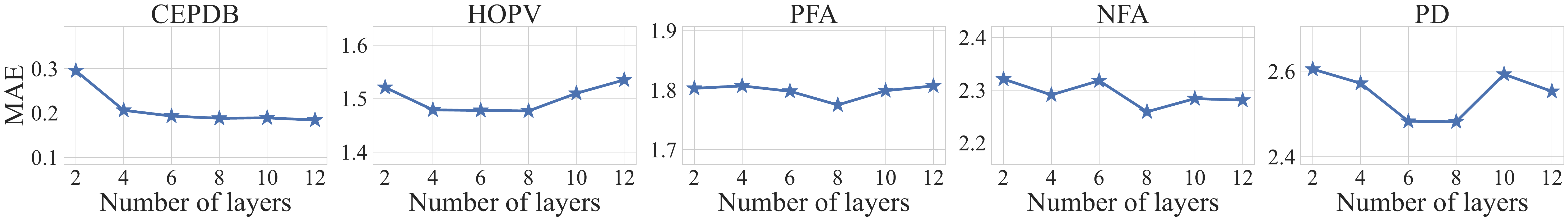}
    \vspace{-2mm}
    \caption{PCE (\%) prediction performance of \algo by test MAE when the number of \algo layers $L$ varies.}
    \label{fig:layer} 
    
\end{figure*}

\header\textbf{Performance under different GNN backbones.}
In the previous experiments, we regard  \gine as the default GNN backbone used in \atom and \inter components of \algo. In this experiment, we assess the performance of \algo using various GNN backbones and provide the results in Table~\ref{tab:backbone}. Specifically, we implement \algo with \gine~\cite{hu2019strategies}, GraphSAGE~\cite{hamilton2017inductive}, and GatedGCN~\cite{bresson2017residual}. To provide a comprehensive comparison, we evaluate the performance of \algo with different GNN backbones against that of the best-performing deep learning-based competitor in Table~\ref{tab:PCE}.
The results in Table~\ref{tab:backbone} reveal that \algo consistently outperforms the best deep learning-based competitor in 4 out of 5 datasets (\cepdb, \polymer, \nnfa, and \pnfa), regardless of the GNN backbone used. This demonstrates the robustness of \algo to variations in GNN backbones. Additionally, we observe that \algo with \gine achieves the best performance more frequently across different datasets. This superiority is attributed to \gine's effective utilization of edge attributes.

\header\textbf{Effect of the number of layers $L$.}
We performed experiments to investigate the impact of the number of layers $L$ in \algo. We vary the range of $L$ from 2 to 12 and assess \algo's performance in PCE prediction across all five datasets using test MAE.
As depicted in Figure~\ref{fig:layer}, we observe an initial decrease in MAE as $L$ increases from 2 to 8 across all datasets. However, with a further increase in $L$, we notice a decline in \algo's performance in \hopv, \polymer, \nnfa, and \pnfa. In contrast, MAE continues to decrease in \cepdb. This observation could be attributed to the fact that the four experimental datasets contain much fewer data  (refer to Table~\ref{tab:dataset} for dataset statistics), making it hard to train deeper models.